\documentclass[final,3p,times,twocolumn]{elsarticle}

\usepackage{amssymb}
\usepackage{amsmath}
\usepackage{amsthm}
\usepackage{algorithmic}
\usepackage{algorithm}
\usepackage{hyperref}

\usepackage{lineno}

\begin{document}

\begin{frontmatter}


\title{Statistical Analysis of the Impact of Quaternion Components in Convolutional Neural Networks} 

\author[1]{Gerardo Altamirano-Gómez\corref{cor1}}
\ead{gerardo.altamirano@iimas.unam.mx}

\author[1,2]{Carlos Gershenson}
\ead{cgg@unam.mx}

\affiliation[1]{organization={Universidad Nacional Autónoma de México. Instituto de Investigaciones en Matemáticas Aplicadas y Sistemas},
            addressline={Circuito Escolar S/N, Ciudad Universitaria},
            city={Coyoacán},
            postcode={04510},
            state={Ciudad de México},
            country={México}}

\affiliation[2]{organization={Binghamton University. Thomas J. Watson College of Engineering and Applied Science},
            addressline={4400 Vestal Parkway East},
            city={Binghamton},
            postcode={13902},
            state={New York},
            country={USA}}

\cortext[cor1]{Corresponding author}

\begin{abstract}
In recent years, several models using Quaternion-Valued Convolutional Neural Networks (QCNNs) for different problems have been proposed. Although the definition of the quaternion convolution layer is the same, there are different adaptations of other atomic components to the quaternion domain, e.g., pooling layers, activation functions, fully connected layers, etc. However, the effect of selecting a specific type of these components and the way in which their interactions affect the performance of the model still unclear. Understanding the impact of these choices on model performance is vital for effectively utilizing QCNNs. This paper presents a statistical analysis carried out on experimental data to compare the performance of existing components for the image classification problem. In addition, we introduce a novel Fully Quaternion ReLU activation function, which exploits the unique properties of quaternion algebra to improve model performance.
\end{abstract}

\begin{graphicalabstract}
\includegraphics[width=\textwidth]{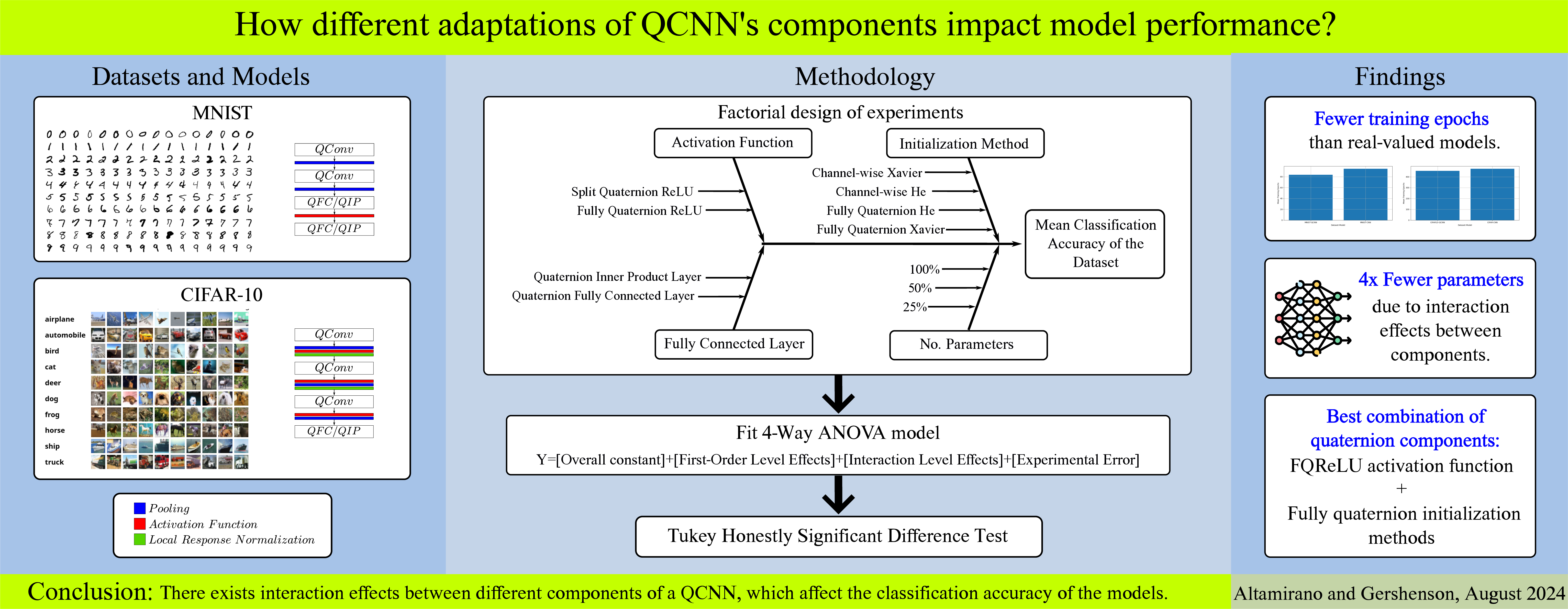}
\end{graphicalabstract}

\begin{highlights}
\item There exists interaction effects between different components of a Quaternion-valued CNN, which affect the classification accuracy of the models.
\item TThe use of the fully quaternion ReLU activation function and fully quaternion initialization methods improves the classification accuracy of Quaternion-valued CNN models.
\end{highlights}


\end{frontmatter}

\section{Introduction}

Among different types of artificial neural networks, those using a combination of convolution and pooling layers have achieved the best performance for image classification tasks~\cite{lecun:1989:mnist, lecun:1998:mnist, krizhevsky:2012:alexnet,simoyan:2014:vggnet, szegedy:2015:googlenet}. A convolution layer is a variation of a perceptron neuron, which applies a weight-sharing technique, similar to the receptive fields discovered by Hubel and Wiesel~\cite{fukushima:1980:neocognitron, hubelwiesel:1968:receptivefields}. This layer, in combination with a pooling layer, produces an invariant signature to a group of geometric transformations~\cite{poggio:2012:computationalmagic, poggio:2014:itheoryproves, poggio:2015:itheory}, e.g. small translations or rotation of the input images~\cite{lecun:1989:designstrategies, lecun:1998:cnn}.

Some of the main problems in designing deep models of CNNs are: reducing the number of parameters without losing generalization, and the vanishing and exploding gradient problems when training the network~\cite{hochreiter:1997:vanishinggradient, pascanu:2013:vanishinggradient}. Fundamental research and experimental analysis have shown that some algebraic systems, such as complex and hyper-complex numbers, have the potential to solve these problems~\cite{nitta:2002:criticalpoints, hirose:2012:generalization, arjovsky:2016:complexrnn, shahadat:2024:hcnn} . Thus, in recent years, several convolutional neural network models using a quaternion representation (QCNNs), instead of the real number representation, have been proposed, see for example~\cite{altamirano:2017:geometricperception, gaudet:2018:qcnn, parcollet:2019:qcnn, zhu:2018:qcnn, grassucci:2021:qcnn,grassucci:2022:qsngan}.

These models have shown that they can achieve similar or better results than their real-valued counterparts. However, the atomic components of each of the quaternion models differ, e.g. activation functions, initialization algorithm, pooling method, etc. This makes experimental data from previous works unsuitable to draw conclusions about the effect of each individual component. In addition, the models are tested over problems of different domains, e.g. image classification~\cite{gaudet:2018:qcnn, zhu:2018:qcnn}, artificial image generation~\cite{grassucci:2021:qcnn,grassucci:2022:qsngan}, natural language processing~\cite{parcollet:2018:qcnn}, etc. and have used different datasets.

The main contributions of this work are as follows:
\begin{itemize}
     \item We present an n-way ANOVA test carried out on experimental data for comparison of the different components of QCNNs for the image classification problem. We selected four factors to test: the type of activation function (Fully Quaternion ReLU or Split Quaternion ReLU function~\cite{gaudet:2018:qcnn, hongo:2020:qcnn, yin:2019:qcnn}), the type of fully connected layer (Quaternion Fully Connected layer~\cite{altamirano:2017:geometricperception} or Quaternion Inner Product layer), the initialization algorithm (channel-wise algorithms~\cite{glorot:2010:weightinicialization, he:2015:weightinicialization} or fully quaternion algorithms~\cite{gaudet:2018:qcnn, parcollet:2019:qcnnrecurrent}) and the number of parameters of the model. We meassure the interaction effect of the factors on the output variable, i.e. classification accuracy, and obtained the combination of factors with best performance, as well as the performance of individual components.
     \item We propose a novel Fully Quaternion ReLU activation function which outperforms the classifcation accuracy achieved by the Split Quaternion ReLU function~\cite{gaudet:2018:qcnn, hongo:2020:qcnn, yin:2019:qcnn}.
\end{itemize}

This paper is organized as follows. Section \ref{sec:methods} introduces definitions and operations related to quaternions and describes the atomic components used for implementing QCNN architectures, including our novel fully quaternion activation function.  Section \ref{sec:experiments} presents the experimental analysis, followed by a discussion at Section \ref{sec:discussion}. Finally, Section \ref{sec:conclusions} states the conclusions and future works.

\section{Methods} \label{sec:methods}
\subsection{Quaternion algebra} \label{sec:qalgebra}

In mid-XIX century, W.R. Hamilton (1805-1865) introduced the concept of quaternion, which he defined as the ratio between two vectors~\cite{hamilton:1853:quaternions, hamilton:1866:quaternions, hamilton:2000:quaternions}. From this definition, he obtained the quadrinomial form of a quaternion~\cite{hamilton:1866:quaternions}:
\begin{equation}
    \textbf{q} = q_R+ q_I\hat{i}+ q_J\hat{j} + q_K\hat{k},
    \label{eq:quaternion}
\end{equation}
where $q_R, q_I, q_J, q_K$ are scalars, and $\hat{i}, \hat{j}, \hat{k}$ are imaginary bases, i.e. $\hat{i}^2=\hat{j}^2=\hat{k}^2=-1$.

In terms of modern mathematics, the quaternion algebra, $\mathbb{H}$, is: the 4-dimensional vector space over the field of the real numbers, generated by the basis $\{1,\hat{i},\hat{j},\hat{k}\}$, and endowed with the following multiplication rules:
\begin{align}
    (1)(1)      &= 1                               \nonumber \\ 
    (1)(\hat{i})&= \hat{j}\hat{k} = -\hat{k}\hat{j} = \hat{i} \nonumber \\
    (1)(\hat{j})&= \hat{k}\hat{i} = -\hat{i}\hat{k} = \hat{j} \nonumber \\
    (1)(\hat{k})&= \hat{i}\hat{j} = -\hat{j}\hat{i} = \hat{k} \nonumber \\
    \hat{i}^2   &= \hat{j}^2      = \hat{k}^2       = -1
\end{align}
The quaternion algebra is associative and non-commutative, and sum between elements and multiplication by a scalar is defined as usual.

Finally, a useful operation when working with quaternions is the \textit{conjugate}. Let, $\textbf{q} = q_R+ q_I\hat{i}+ q_J\hat{j} + q_K\hat{k}$, be a quaternion, its conjugate, $\bar{\textbf{q}}$ , is defined as follows:
\begin{equation}
    \bar{\textbf{q}} = q_R - q_I\hat{i} - q_J\hat{j} - q_K\hat{k}.
\end{equation}

\subsection{QCNNs components} \label{sec:qcnncomponets}

The first step when working with QCCNs is to map the data to the quaternion domain. For a dataset containing images, an RGB image is mapped to the quaternion domain by encoding the red, green, and blue channels into the imaginary parts of the quaternion:
\begin{equation}
    \mathbf{q}=0+R\hat{i}+G\hat{j}+B\hat{k},
\end{equation}

In contrast, for grayscale images, the grayscale values are mapped to the real part of the quaternion, and the imaginary components are set to zero. 

Next, we introduce the main atomic components for designing QCNNs architectures.

\subsubsection{Quaternion convolution layers}

Each sample, denoted by $\mathbf{Q}$, is represented as an $N\times M$ matrix where each element is a quaternion:
\begin{equation}
    \mathbf{Q}= [\mathbf{q}(x,y)] \in \mathbb{H}^{N\times M};
\end{equation}
in addition,  $\mathbf{Q}$ can be decomposed in its real and imaginary components:
\begin{equation}
    \mathbf{Q}= {Q_R}+{Q_I}\hat{i}+{Q_J}\hat{j}+{Q_K}\hat{k}
\end{equation}
where ${Q_R}, {Q_I}, {Q_J}, {Q_K} \in \mathbb{R}^{N\times M}$, and $\hat{i},\hat{j},\hat{k}$ represent the complex basis of the quaternion algebra.

In the same way, a convolution kernel of size $L\times L$ is represented by a quaternion matrix, as follows:
\begin{equation}
    \mathbf{W}= [\mathbf{w}(x,y)] \in \mathbb{H}^{L\times L}
\end{equation}
which can be decomposed as:
\begin{equation}
    \mathbf{W}= {W_R}+{W_I}\hat{i}+{W_J}\hat{j}+{W_K}\hat{k},
\end{equation}
where ${W_R},{W_I}, {W_J}, {W_K} \in \mathbb{R}^{L\times L}$, and $\hat{i},\hat{j},\hat{k}$ represent the basis of the quaternion algebra.

Using the left-sided definition of discrete quaternion convolution, the convolution layer is defined as follows:
\begin{equation}\label{eq:qcnn:qconv1}
    \mathbf{F} = \mathbf{W} \ast \mathbf{Q},
\end{equation}
where $\mathbf{F}\in \mathbb{H}^{(N-L+1)\times (M-L+1)}$ represents the output of the layer, i.e. a quaternion feature map, and each element of the tensor is computed as follows~\cite{ell:1993:qconv, ell:2007:qconv, pei:2001:qft}:
\begin{equation}
    \mathbf{f}(x,y) = (\mathbf{w} \ast \mathbf{q})(x,y).
    \sum_{r=-\frac{L}{2}}^{\frac{L}{2}} \sum_{s=-\frac{L}{2}}^{\frac{L}{2}} [\mathbf{w}(r,s) \mathbf{q}(x-r,y-s)].
\end{equation}

In addition, for intermediate layers with more than four channels, the input data is divided into 4-channel sub-inputs. Each sub-input is convoluted with a different quaternion kernel to produce a partial output, and the final quaternion feature map, is computed by summing all the partial outputs. This is stated formally as follows: Let $\mathbf{X} \in \mathbb{R}^{N\times M \times C}$ be an input data, $N$ is the number of rows, $M$ the number of columns, and $C$ is the number of channels, where $C\%4=0$, then $X$ is partitioned as follows:
\begin{equation}
    \mathbf{X}=  [\mathbf{Q_0}, \mathbf{Q_1}, \dots, \mathbf{Q_{(C/4)-1}} ]
\end{equation}
where each $\mathbf{Q_s} \in \mathbb{H}^{N\times M}$, $0<s<(C/4)-1$ is a \textit{quaternion input channel}.

Let $\mathbf{V} \in \mathbb{R}^{L\times L \times K}$, with $K\%4=0$, be the convolution kernel, then:
\begin{equation}
    \mathbf{V}=  [\mathbf{W_0}, \mathbf{W_1}, \dots, \mathbf{W_{(K/4)-1}} ]
\end{equation}
where each $\mathbf{W_s} \in \mathbb{H}^{L\times L}$, $0<s<(K/4)-1$ is a \textit{quaternion kernel channel}.

Then, each partial output is computed as follows:
\begin{equation}
    \mathbf{F_s}= \mathbf{W_s} * \mathbf{Q_s},
\end{equation}
where $0<s<(C/4)-1$, and the final quaternion feature map, $\mathbf{F}\in \mathbb{H}^{N\times M}$, is obtained by summing all outputs:
\begin{equation}
    \mathbf{F}  =  \sum_{s=0}^{(C/4)-1} \mathbf{F_s}.
\end{equation}

\subsubsection{Pooling layers}

A pooling layer introduces a sort of invariance to geometric transformations, such as small translations and rotations. In this work, it is applied a channel-wise average layer~\cite{lin:2013:networkinnetwork, lin:2013:globalpooling, zhu:2018:qcnn} as follows: First, we select a window from the input image, denoted by $\mathbf{Q} \in \mathbb{H}^{L_1\times L_2}$; then, the pooling procedure is applied on this sub-image:
\begin{equation}
    SQAvePool(\mathbf{Q}) = \frac{1}{L_1 L_2}\sum_{i=1}^{L_1}\sum_{j=1}^{L_2}q(i,j)
\end{equation}
This process is repeated over the whole image by moving the window mask from point to point in the input image.

\subsubsection{Activation functions}

The role of an activation function is to simulate the triggering behavior of a biological neuron. From the computational perspective,  non-linear activation functions are required for constructing a universal interpolator of a continuous quaternion valued function~\cite{arena:1997:qmlp2}. In addition, it is desirable that the activation function were analytic so that gradient descendant techniques can be applied in the training stage. However, the non-analytic condition can only be satisfied by some linear and constant functions~\cite{deavours:1973:qcalculus, sudbery:1979:cauchyriemannfuetereq}. A typical way to circumvent this problem is the use of quaternion splits functions, i.e. a mapping $f:\mathbb{H}\rightarrow \mathbb{H}$, such that:
\begin{equation}
    f(\mathbf{q})=f_R(\mathbf{q})+f_I(\mathbf{q})\hat{i}+f_J(\mathbf{q})\hat{j}+f_K(\mathbf{q})\hat{k},
\end{equation}
where $\mathbf{q}\in\mathbb{H}$, $f_R$, $f_I$, $f_J$, and $f_K$  are mappings over the real numbers: $f_{*}:\mathbb{R}\rightarrow \mathbb{R}$. Thus, the Split Quaternion ReLU activation function~\cite{gaudet:2018:qcnn, hongo:2020:qcnn, yin:2019:qcnn} is defined as follows:
\begin{eqnarray}
    SQReLU(\mathbf{q}) &=& ReLU(q_R) + ReLU(q_I)\hat{i} + \\
    && ReLU(q_J)\hat{j} + ReLU(q_K)\hat{k},
\end{eqnarray}
where $ReLU: \mathbb{R}\rightarrow \mathbb{R}$ is the real-valued ReLU function~\cite{fukushima:1969:relu, goodfellow:2016:deeplearning}:
\begin{equation}
    ReLU(x)= max(0, x).
\end{equation}

A more general approach is the use of Fully Quaternion Functions, i.e. a mapping $f:\mathbb{H}\rightarrow \mathbb{H}$. Based on the Complex ReLU activation function~\cite{guberman:2016:c2n2,trabelsi:2018:complexnn}, we propose the Fully Quaternion ReLU activation function, defined as follows:
\begin{equation}
    FQReLU(\mathbf{q})=
    \begin{cases}
        \mathbf{q} & \text{if } \theta \in [0,\pi/2] \\
        0 & \text{otherwise}
    \end{cases}
\end{equation}
 where $\mathbf{q}=q_R+q_I\hat{i}+q_J\hat{j}+q_K\hat{k} \in \mathbb{H}$, and $\theta$ is the phase of the quaternion, computed as follows~\cite{ward:1997:quaternionscayley}:
 \begin{equation}
    \theta = atan\left(\frac{\sqrt{q_I^2+q_J^2+q_K^2} }{q_R}\right)
 \end{equation}

For these functions, the learning dynamics are built using partial derivatives on the quaternion domain.

\subsubsection{Fully Connected Layers}

Let $\mathbf{Q}$ be a $N_1\times N_2 \times N_3$ tensor, representing the input to a fully connected layer; then each element of $\mathbf{Q}$ is a quaternion:
\begin{equation}
    \mathbf{Q}= [\mathbf{q}(x,y,z)] \in \mathbb{H}^{N_1\times N_2 \times N_3},
\end{equation}
where $N_1, N_2, N_3$ are the height, width and number of channels of the input. 

Now, it is defined a quaternion kernel, $\mathbf{W}$, of size $N_1\times N_2 \times N_3$, where each element is a quaternion:
\begin{equation}
    \mathbf{W}= [\mathbf{w}(x,y,z)] \in \mathbb{H}^{N_1\times N_2 \times N_3}.
\end{equation}
where $N_1, N_2, N_3$ are the height, width and number of channels of the input. Note that elements of the input and weight tensors are denoted as $q(x,y,z)$ and $w(x,y,z)$, respectively.

Thus, for Quaternion Fully Connected layers (QFC), the output, $\mathbf{f}$, is computed as follows~\cite{altamirano:2017:geometricperception}:
 \begin{equation}
    \mathbf{f} = \sum_{r,s,t}^{N_1,N_2,N_3} \left[\mathbf{w}(r,s,t) \mathbf{q}(r,s,t)\right].
\end{equation}
hence, $\mathbf{f}$ is a single quaternion. 

Alternatively, we have the Quaternion Inner Product layer (QIP); in this case, the output, $\mathbf{f}$, is computed as follows:
 \begin{equation}
    \mathbf{f} = \sum_{r,s,t}^{N_1,N_2,N_3} \left[\mathbf{w}*(r,s,t)\cdot \mathbf{q}(r,s,t)\right].
\end{equation}
where $\cdot$ is the inner product; hence, $\mathbf{f}$ is a single real number. Notice that this type of layer produces the same output that a multichannel real-valued fully connected or real-valued inner product layers. In contrast, the output of a quaternion fully connected layer differs from a quaternion inner product layer, since the former one is a quaternion value while the later is a real value.

\subsubsection{Initialization methods}

Weight initialization algorithms rely on the idea of making the variance dependent of each layer, so activation and back-propagated gradient variances are maintained as we move up or down the network. In this way, the vanishing and exploding gradients problems are mitigated~\cite{hochreiter:1997:vanishinggradient, pascanu:2013:vanishinggradient}.

In the quaternion domain, Gaudet and Maida~\cite{gaudet:2018:qcnn} as well as Parcollet \textit{et al.}~\cite{parcollet:2019:qcnnrecurrent} extended available methods in the real-valued domain. These works consider each quaternion weight as a 4-dimensional vector, which components are independent, normally distributed, centered at zero. Then, the weight initialization relies on selecting the mode of a 4DOF Rayleigh distribution, denoted by $\sigma$. If $\sigma={1}/{\sqrt{2(n_{in}+n_{out})}}$ we have a \textit{quaternion normalized initialization} (or QXavier initialization) which ensures that the variances of the quaternion input, output and their gradients are the same; while $\sigma={1}/{\sqrt{2n_{in}}}$ is used for the \textit{quaternion ReLU initialization} (or QHe initialization), where  $n_{in}$, and $n_{out}$ are the number of neurons of the input and output layers, respectively. The initialization method in algorithmic form is presented in Algorithm \ref{alg:weightinitialization}.


\begin{algorithm}[!t]
  \caption {Quaternion-valued weight initialization} \label{alg:weightinitialization}
  \begin{algorithmic}[1]
    \REQUIRE $\mathbf{W}\in\mathbb{H}^{n_{in}\cdot n_{out}}, n_{in}\in\mathbb{N}^+, n_{out}\in\mathbb{N}^+$
    \IF {RELU} \STATE {$\sigma\gets\frac{1}{\sqrt{2n_{in}}}$} \ELSE \STATE{$\sigma\gets\frac{1}{\sqrt{2(n_{in}+n_{out})}}$} \ENDIF
    \FORALL{$\mathbf{w}$ in $\mathbf{W}$}
      \STATE {$\phi\gets rayleight\_rand(\sigma)$}
      \STATE {$\theta\gets uniform\_rand(-\pi,\pi)$} 
      \STATE {$x,y,z\gets uniform\_rand(0,1)$}
      \STATE {$\hat{u}\gets \frac{x\hat{i}+y\hat{j}+z\hat{k}}{\sqrt{x^2+y^2+z^2}}$}
      \STATE {$\mathbf{w}\gets \phi\cos(\theta)+\phi\sin(\theta)\hat{u}$}
    \ENDFOR\\
    \RETURN $\mathbf{W}$
  \end{algorithmic}
\end{algorithm}

In addition to these algorithms, in our experiments, we tested another two methods applied in a channel-wise manner: the Glorot's normalized initialization (or Xavier initialization)
~\cite{glorot:2010:weightinicialization}, and the He's algorithm for non-Linear activation functions (or He initialization)~\cite{he:2015:weightinicialization}.

\subsubsection{Training}

The proposed models were trained using the Quaternion Backpropagation Algorithm~\cite{altamirano:2017:geometricperception, arena:1994:qmlp0, arena:1996:qmlp1, arena:1998:qbackprop}, where derivatives are computed using the Generalized Quaternion Chain Rule for a Real-valued function~\cite{gaudet:2018:qcnn, sudbery:1979:cauchyriemannfuetereq}. The algorithm is summarized as follows:

Let $\mathbf{Q} = [\mathbf{q}(x,y)] \in \mathbb{H}^{N\times M}$ be the input to a convolution layer, $\mathbf{W} = [\mathbf{w}(x,y)] \in \mathbb{H}^{L\times L}$ be the weights of the convolution kernel, and $\mathbf{F} = [\mathbf{f}(x,y)]\in \mathbb{H}^{(N-L+1)\times (M-L+1)}$ be the output of the layer; then, a quaternion weight, $\mathbf{w}(s,t)$, where $1<s,t<L$, represents the weight connecting input $\mathbf{q}(a+s,b+t)$ with output $\mathbf{f}(u,v)$, where $1<u<N-L+1$ and $1<v<N-L+1$. In addition, let $\mathbf{d}(u,v)^{top}, \mathbf{d}(s,t)^{bottom} \in\mathbb{H}$ be the error propagated from the top and to the bottom layers, respectively, and $\epsilon\in\mathbb{R}^+$ be the learning rate; then, for the current convolution layer:

\begin{enumerate}
    \item Update its weights using the following equation:
    \begin{equation} \label{eq:backprop_w}
        \mathbf{w}(s,t)=\mathbf{w}(s,t) + \epsilon \mathbf{d}(u,v)^{\text{top}} \mathbf{\bar{x}}(u+s,v+t)
    \end{equation}
    \item Update the bias term:
    \begin{equation} \label{eq:backprop_b}
      \mathbf{b}(u,v)=\mathbf{b}(u,v) + \epsilon  \mathbf{d}(u,v)^{\text{top}},
    \end{equation}
    \item Propagate the error to the bottom layer according to the following equation:
    \begin{equation} \label{eq:backprop_delta}
      \mathbf{d}(s,t)^{\text{bottom}} =  \sum_{u}\sum_{v} (\mathbf{\bar{w}}(s,t) \mathbf{d}(u,v)^{\text{top}})
    \end{equation}
\end{enumerate}

Note that Equations (\ref{eq:backprop_w}) and (\ref{eq:backprop_delta}) apply quaternion products. 

For an activation layer, the error is propagated to the bottom layer according to the following equation:
\begin{equation}  \label{eq:backprop_actfn}
  \mathbf{d}(s,t)^{\text{bottom}} =   \mathbf{d}(u,v)^{\text{top}} \odot f'(\mathbf{x}(u+s,v+t))
\end{equation}
where $\odot$ is the component wise product.

\subsection{Factorial Design of Experiments.}
\begin{figure*}[!t]
    \centering
    \includegraphics[width=0.9\textwidth]{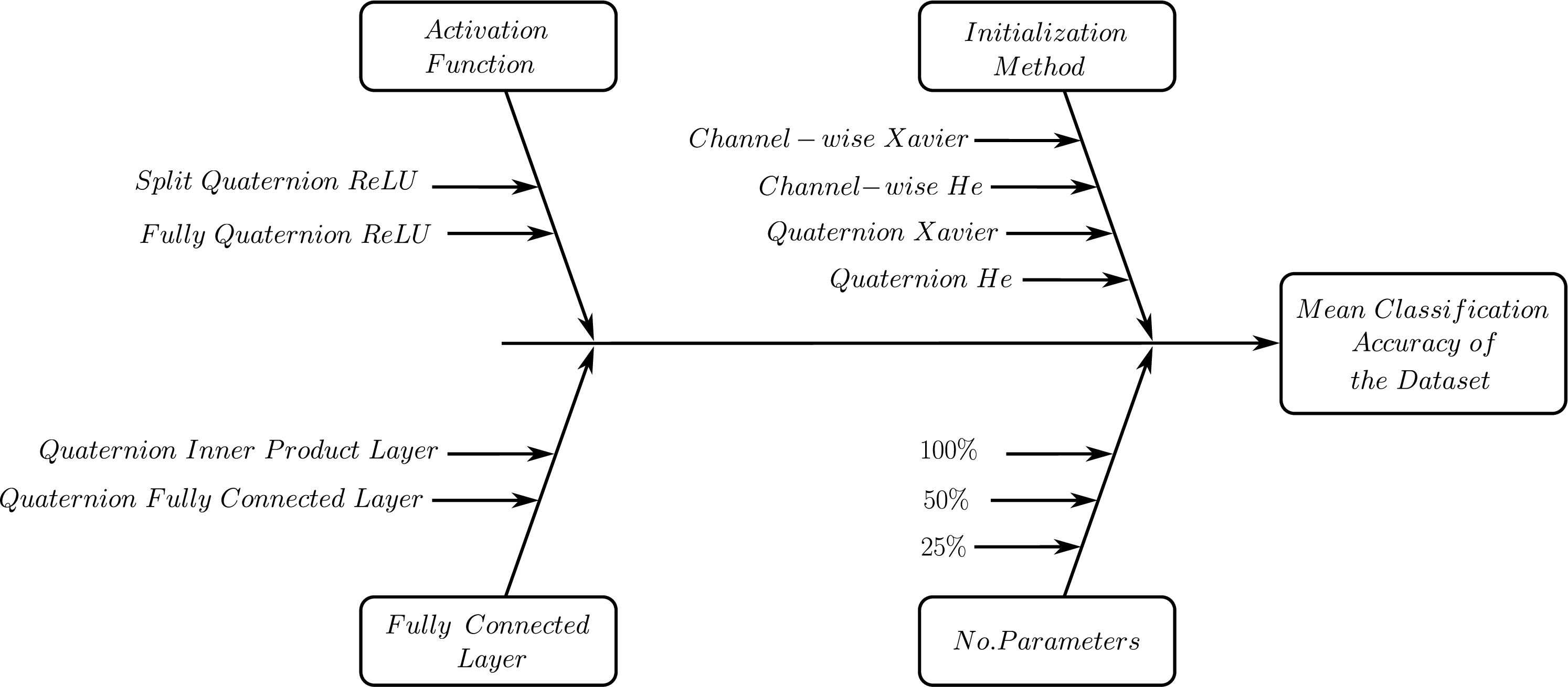} 
    \caption {Cause-and-effect diagram for the statistical design of experiments.} \label{fig:ishakawa}
\end{figure*}

In order to obtain an effective statistical analysis, a proper statistical design of the scientific study must be carried on. For deep learning models, we would like to demonstrate the cause-and-effect relationship between some explanatory \textit{factors}, i.e. type of activation function, type of weight initialization method, type of fully connected layer and number of parameters; and the \textit{response variable}, i.e. the mean classification accuracy obtained for the complete dataset at the training or testing stages. The cause-and-effect relationship is demonstrated by changing the \textit{levels} or \textit{treatments} of the factors and observing their effect on the response variable. For example, for the activation function factor we can select between the fully quaternion or the split quaternion function. For the factors proposed in this study, all levels can be set by the investigator, thus we are dealing with \textit{experimental factors}, and all factors but the number of parameters are qualitative. Moreover, all factors are investigated simultaneously; thus, we have a crossed multifactor study also called a \textit{factorial design} \cite{kutner:2005:statistics}. Figure \ref{fig:ishakawa} shows the cause-and-effect diagram of the study, containing the four potential factors and their levels, leading to a total of $24$ level combinations. 

A particular deep learning model to which a combination of levels can be applied is called the \textit{experimental unit}. If each combination of levels is applied more than once, then it is said that we have \textit{complete replicates} of the experiment, which help to estimate the experimental error variance, the presence of level effects and the confidence intervals of the estimations.

In the design of the experiments, the assignment of levels to the experimental unit can be achieved by a randomization procedure \cite{kutner:2005:statistics}; however, for this study we decided to run all possible combinations of levels in order to obtain a full dataset. Thus, for analyzing the effect of the factors on the model, a set of levels are assigned to the deep learning model, it is trained, and the mean classification accuracy for the complete dataset is computed. The procedure is repeated for each possible combination of levels with $10$ complete replicates. In this way, we obtained a full dataset containing samples of the responses for all possible combinations of levels.

Corresponding to each combination of factor levels, there is a probability distribution of responses. If the sample mean of each distribution is computed, it is very likely that they would differ; however, we must know if this difference is statistically significant if we want to establish a cause-effect relationship between factors and response. The analysis of variance is the statistical procedure that analyze the difference in the sample means, so we can state if the probability distribution means are different with some degree of certainty \cite{ott:2010:statistics}. Thus, the factorial design of experiments is associated with a linear statistical model, which has the following general form \cite{kutner:2005:statistics}:
\begin{eqnarray}
    Y &=& [\text{Overall constant}] \\ \nonumber
    && + [\text{First-Order Level Effects}] \\ \nonumber
    && + [\text{Interaction Level Effects}] \\ \nonumber
    && + [\text{Experimental Error}]
\end{eqnarray}
Note that this model incorporates the effect of the factors on the response variable as well as the interaction effects among the individual factors.
In addition, the analysis of variance takes the following assumptions \cite{kutner:2005:statistics}: 
\begin{itemize}
     \item The probability distribution of each response is normal.
     \item The probability distribution of each response has the same variance.
     \item The responses for each factor level are random selections from the corresponding probability distribution and are independent of the responses for any other factor.
\end{itemize}
Due to these constrains, the only possible difference between the probability distributions is their means; thus, the analysis of variance focuses on analyzing the mean responses for determining the effect of the different factor levels on the responses. If the factor level means are equal, then there is no relationship between the factor and the responses; however, a relationship between factors and response exists if they differ, and further analysis of the factor levels is required. Such analysis is carried out by the Tukey test, which test each factor level with every other one to determine the statistically significant set of comparisons, as well as estimates two-sided confidence intervals. This analysis determines the best combination of levels for all factors. In addition, a paired comparison plot can be computed for visual analysis of results.

Finally, the statistical procedure can be summarized as follows:
\begin{enumerate}
     \item Set-up the factorial design of experiments considering 4 factors: type of activation function, type of weight initialization method, type of fully connected layer, number of parameters; and the response variables: mean classification accuracy obtained for the complete dataset at the training or testing stages.
     \item Fit a 4-way ANOVA model, and reduce the model if no interaction effects are present.
     \item Check the ANOVA's model assumptions: normality and equality of variances.
     \item Perform Tukey's multiple comparison procedure and determine the best combination of levels.
\end{enumerate}

\section{Experimental analysis}\label{sec:experiments}

This work is focused in analyzing the behavior of different QCNNs components; thus, we intentionally used simple architectures for classifying images from classic datasets, such as MNIST~\cite{lecun:1989:mnist} and CIFAR10~\cite{krizhevsky:2009:cifar}. All models were programmed, with GPU support, on a modified version of Caffe \cite{jia:2014:caffe}; experiments were carried on a server with $12$ CPUs Intel Xeon E5-2640, 64Gb of RAM, Graphics Card Nvidia Tesla K20m and running the Ubuntu Server 18.04 Operative System.

\subsection{MNIST dataset}
\begin{figure}[!t]
    \centering
    \includegraphics[width=0.275\textwidth]{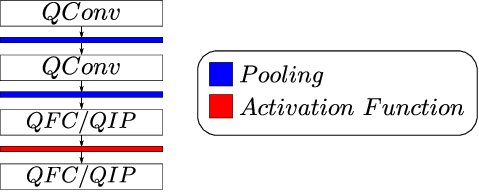} 
    \caption {Architecture used for classification of the MNIST dataset.} \label{fig:mnistmodels}
\end{figure}

In this set of experiments, we used the architecture presented in Figure \ref{fig:mnistmodels}, where, for each model, we tested different types of activation functions: Split Quaternion ReLU (SQReLU) vs. Fully Quaternion ReLU (FQReLU); fully connected layers: Quaternion Inner Product (QIP) vs. Quaternion Fully Connected (QFC); and initialization methods: Glorot's method applied in a channel-wise manner (Xavier)~\cite{glorot:2010:weightinicialization}, He's method applied in a channel-wise manner (He)~\cite{he:2015:weightinicialization}, quaternion normalized initialization (QXavier) and quaternion ReLU initialization (QHe)~\cite{gaudet:2018:qcnn, parcollet:2019:qcnnrecurrent}. Moreover, we tested each model with different number of parameters. Table ~\ref{tab:qmnistmodels} shows the full list of models as well as their parameters for each layer.

\begin{table*}
    \centering
    \begin{tabular}{|c|c|c|c|c|c|}
    \hline
    {Name} & {CONV-1} & {CONV-2} & {FC-1} & {FC-2} & {No. Parameters} \\
    \hline
    FQReLU-QIP-1 & $10$ & $25$ & $256$ & $10$ & $438,160$ \\
    FQReLU-QIP-1-2 & $5$ & $13$ & $128$ & $10$ & $114,776$ \\
    FQReLU-QIP-1-4 & $3$ & $6$ & $64$ & $10$ & $27,316$ \\
    \hline
    FQReLU-QFC-1 & $20$ & $50$ & $128$ & $3$ & $513,136$ \\
    FQReLU-QFC-1-2 & $10$ & $25$ & $64$ & $3$ & $129,168$ \\
    FQReLU-QFC-1-4 & $5$ & $13$ & $32$ & $3$ & $34,008$ \\
    \hline
    SQReLU-QIP-1 & $10$ & $25$ & $256$ & $10$ & $438,160$ \\
    SQReLU-QIP-1-2 & $5$ & $13$ & $128$ & $10$ & $114,776$ \\
    SQReLU-QIP-1-4 & $3$ & $6$ & $64$ & $10$ & $27,316$\\
    \hline
    SQReLU-QFC-1 & $20$ & $50$ & $128$ & $3$ & $513,136$ \\
    SQReLU-QFC-1-2 & $10$ & $25$ & $64$ & $3$ & $129,168$ \\
    SQReLU-QFC-1-4 & $5$ & $13$ & $32$ & $3$ & $34,008$ \\
    \hline
    CONV-ReLU-IP & $20$ & $52$ & $500$ & $10$ & $449,000$ \\
     \hline
   \end{tabular}
    \caption{Models tested for classifying images of the MNIST dataset. The first $12$ rows show models that apply the quaternion convolution, while the last row is a model that applies the real-valued convolution (the name starts with CONV). Each model was named as follows: SQRELU indicates that it uses Split quaternion activation functions; FQReLU indicates the use of fully quaternion activation functions, and ReLU indicates the use of real-valued ReLU functions. For fully connected layers: QIP indicates the use of quaternion inner product layers; QFC indicates the use of quaternion fully connected layers, and IP is used for real-valued inner product layers. The columns show the following information: CONV-1 and CONV-2 indicate the number of outputs of each convolution layer (all kernels have sizes $5\times5$), if the output comes from a real-valued convolution layer, the value indicates the number of real-valued outputs, but if the output comes from a quaternion convolution layer, the value indicates the number of quaternion-valued outputs. FC-1 and FC-2 show the number of outputs on each fully connected layer; in the same way, if the output comes from a quaternion-valued or real-valued inner product, the value indicates the number of real-valued outputs, but if the output comes from a quaternion fully connected layer, the value indicates the number of quaternion-valued outputs.} \label{tab:qmnistmodels}
 \end{table*}

Experiments were conducted in the following way: for each model we obtained 10 complete replicates; each replicate consisted in training the model for 100 epochs, information about the training and testing performance were saved at each epoch. Thereafter, for each replicate, we selected the epoch in which the maximum value at the testing stage was obtained. This value, its corresponding training performance value, as well as the epoch was saved for each replicate. The procedure was repeated for each model.

Thereafter, a statistical analysis was conducted using an ANOVA test with 4 factors: initialization method, activation function, type of fully connected layer and number of parameters. Since the overwhelming majority of models obtained $100\%$ of training accuracy, we used the testing accuracy as output variable for comparing the performance between models. Normal population distribution and equality of variances assumptions were checked.

The 4-way ANOVA study concluded that the overall interaction between the four factors was not statistically different ($F=0.7895$, $p-value=0.6897$), i.e. there was not enough evidence supporting that the average accuracy of the groups was different. Thus, we reduced the model to include all possible 3-factor interactions, executed a 3-way ANOVA study, and removed all not statistically different 3-factor interactions. In this way, we obtained a reduced 3-factor model:

\begin{equation}
    Y_{ijkl} = \mu + \alpha_i+\beta_j+\gamma_k+\delta_l
    +(\alpha\gamma\delta)_{ikl}+(\beta\gamma\delta)_{jkl}+\epsilon_{ijkl}
\end{equation}
where $Y_{ijkl}$ are the samples of the testing accuracy; $\mu$ is the grand mean; $\alpha_i=\mu-\mu_i$ is the effect of the initialization method, and $\mu_i$ is the mean of its population; $\beta_j=\mu-\mu_j$ is the effect of the activation function, and $\mu_j$ is the mean of its population; $\gamma_k=\mu-\mu_k$ is the effect of the activation function, and $\gamma_k$ is the mean of its population; $\delta_l=\mu-\mu_l$ is the effect of the activation function, and $\mu_l$ is the mean of its population; $(\alpha\gamma\delta)_{ikl}$ is the interaction effect of the initialization method, the fully connected layer and the number of parameters; $(\beta\gamma\delta)_{jkl}$ is the interaction effect of the activation function, the fully connected layer and the number of parameters; and $\epsilon_{ijkl}$ are the error terms, which are independently normally distributed random variables with expected value of zero.

The ANOVA analysis on this model showed a statistically significant interaction effect on the output variable between: initialization method, fully connected layer, and number of parameters ($F=2.29$, $p-value=3.93\times10^{-3}$); and activation function, fully connected layer, and number of parameters ($F=84.06$, $p-value=1.51\times10^{-61}$). Thus, it is the average response differences among these factors that matters.

\begin{figure*}[!t]
    \centering
    \includegraphics[width=0.485\textwidth]{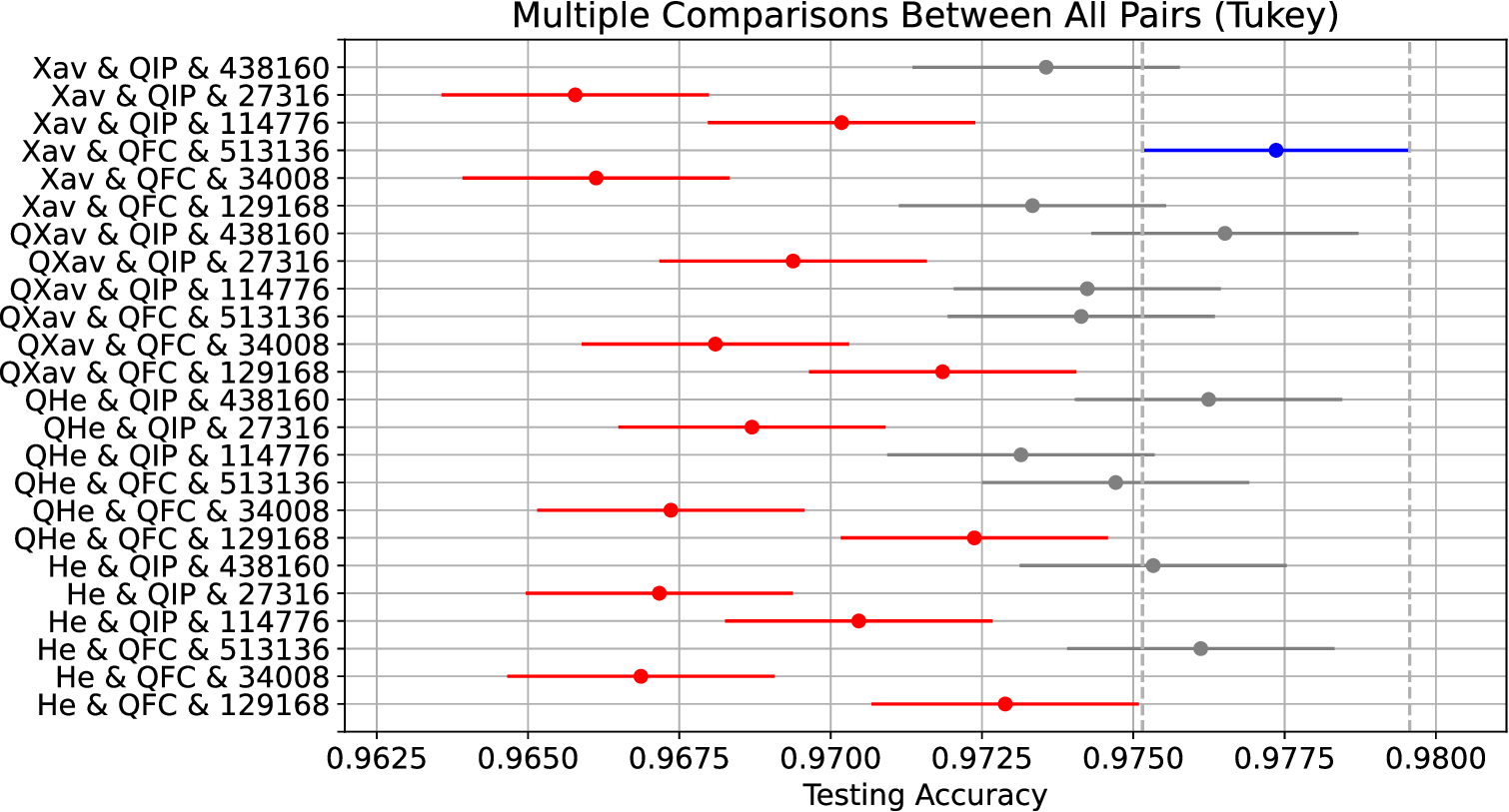} \hfill \includegraphics[width=0.495\textwidth]{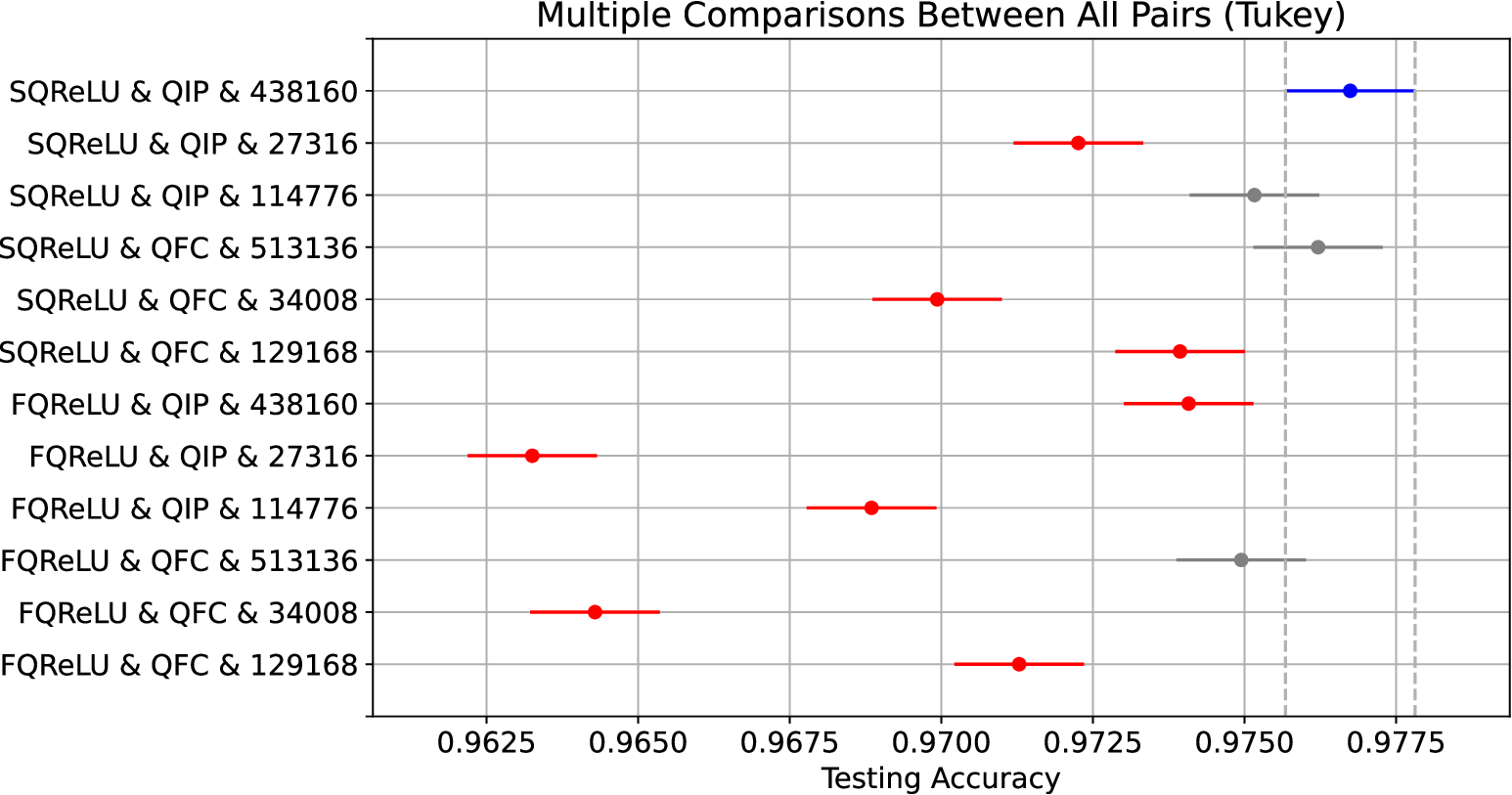}
    \caption {Plot of group means with confidence intervals using data from the MNIST classification task. Left: Combination of initialization method, fully connected layer and number of parameters. Right: Combination of activation function, fully connected layer and number of parameters. The group with higher performance is selected in blue; the groups that are not statistically different from it are shown in gray, while the statistically different groups are shown in red.} \label{fig:mniststats}
\end{figure*}

In order to determine the statistically significant combinations of parameters within each group, we conducted a Tukey Honestly Significant Difference (HSD) test, with $\alpha=0.05$. In this way, we found group combinations of the interaction terms that produce a superior effect on the testing accuracy (output variable). Then, for the analysis of the first interaction group, we pooled the data of the activation function factor, and considered the combination of initialization method, fully connected layer and number of parameters groups; in this case, the best result was produced by the channel-wise Xavier initialization, a QFC layer and the models with $513,136$ parameters (models FQReLU-QFC-1 and SQReLU-QFC-1 in Table~\ref{tab:qmnistmodels}). In addition, their performance was not statistically different from $10$ of the models (p-values included in the supplementary material). The plot of group means with confidence intervals is shown in Figure \ref{fig:mniststats} (left). From this analysis, we concluded that for the models with a higher number of parameters, the initialization method or the type of fully connected layer does not affect their performance; even more some models with $4x$ less parameters, such as QFC-1-2 with Channel-wise Xavier initialization, and QIP-1-2 with Fully Quaternion initialization achieve the same performance. However, there was a statistically significant difference in the performance when reducing the number of parameters by a factor of $8$.

For the second analysis of the interaction terms, we pooled the data of the initialization method and considered the combination of: activation function, fully connected layer and number of parameters groups. The best results were achieved by models SQReLU-QIP-1, SQReLU-QIP-1-2, SQReLU-QFC-1, and FQReLU-QFC-1 (see Table~\ref{tab:qmnistmodels}). The plot of group means with confidence intervals is shown in Figure \ref{fig:mniststats} (right).

In a third analysis, we ran a statistical analysis on individual factors. We found no statistically significant difference between initialization methods (p-values included in the supplementary material) and in fully connected layers ($meandiff=0$, $p-adjusted=0.9$). However, the SQReLU function presented a slightly better performance than the FQReLU function ($meandiff=0.0046$, $p-adjusted=0.001$); this represents less than $0.5\%$ in accuracy percentage). For the number of parameters, the largest models presented a statistically significant difference versus the rest of them (p-values included in the supplementary material). 

These analyses make clear the necessity of considering the interaction effects between components when designing a model.

Thereafter, in a final analysis, we compared the quaternion models with best performance, i.e. FQReLU-QIP-1, SQReLU-QIP-1, FQReLU-QFC-1 and SQReLU-QFC-1 versus a real-valued model containing a similar number of parameters (CONV-ReLU-IP). When we pooled the data of the activation function factor, there was no statistically significant difference in the performance between the real-valued model and the quaternion valued models that used: Xavier initialization and a QFC layer; and QXavier initialization and a QIP layer (p-values included in the supplementary material). However, when we pooled the data of the initialization factor, there was a statistically significant difference in the performance between the real-valued model and all the quaternion valued models (p-values included in the supplementary material). Thus, we concluded that the selection of the initialization method affects in a statistically significant manner the performance of quaternion models. In addition, we noted that the vast majority of quaternion models achieved their best performance in fewer training epochs than the real-valued model. Table~\ref{tab:qmnistperformance} shows the mean and standard deviation values of the training accuracy and the testing accuracy for each model. In addition, it is shown how many epochs it took, on average, to obtain the best performance of the model.

\begin{table*}
  \centering
    \resizebox{\textwidth}{!}{ 
    \begin{tabular}{|c|c|c|c|c|c|c|c|c|c|c|c|c|c|c|c|c|c|c|c|c|c|c|c|c|c|}
    \hline
    & & \multicolumn{6}{c|}{QXavier Initialization} & \multicolumn{6}{c|}{QHe Initialization} & \multicolumn{6}{c|}{Xavier Initialization} & \multicolumn{6}{c|}{He Initialization} \\
    \cline{3-26}
    {Name} & {No.} & \multicolumn{2}{c|}{Epochs} & \multicolumn{2}{c|}{$\%$ Train} & \multicolumn{2}{c|}{$\%$ Test} & \multicolumn{2}{c|}{Epochs} & \multicolumn{2}{c|}{$\%$ Train} & \multicolumn{2}{c|}{$\%$ Test} & \multicolumn{2}{c|}{Epochs} & \multicolumn{2}{c|}{$\%$ Train} & \multicolumn{2}{c|}{$\%$ Test} & \multicolumn{2}{c|}{Epochs} & \multicolumn{2}{c|}{$\%$ Train} & \multicolumn{2}{c|}{$\%$ Test} \\
    \cline{3-26}
    & Param & $\mu$ & $\sigma$ & $\mu$ & $\sigma$ & $\mu$ & $\sigma$ & $\mu$ & $\sigma$ & $\mu$ & $\sigma$ & $\mu$ & $\sigma$ & $\mu$ & $\sigma$ & $\mu$ & $\sigma$ & $\mu$ & $\sigma$ & $\mu$ & $\sigma$ & $\mu$ & $\sigma $ & $\mu$ & $\sigma$ \\
    &  & & & & & & $(10^{-3})$ & & & & & & $(10^{-3})$ & & & & & & $(10^{-3})$ & & & & & & $(10^{-3})$ \\
    \hline

    \textbf{FQReLU-QFC-1} & 513136 & 83 & 13.3 & 1 & 0 & 0.974 & 1.12 & 80 & 15.9 & 1 & 0 & 0.973 & 1.95 & \textbf{91} & \textbf{7.8} & \textbf{1} & \textbf{0} & \textbf{0.977} & \textbf{0.93} & {81} & {16.4} & {1} & {0} & {0.976} & {1.54}\\
    \textbf{SQReLU-QFC-1} & 513136 & 82 & 19.9 & 1 & 0 & 0.974 & 3.25 & {78} & {18.8} & {1} & {0} & {0.976} & {3.22} & \textbf{85} & \textbf{16.5} & \textbf{1} & \textbf{0} & \textbf{0.978} & \textbf{1.63} & {77} & {22.1} & {1} & {0} & {0.976} & {2.53}\\
    FQReLU-QIP-1 & 438160 & 94 & 5.7 & 1 & 0 & 0.975 & 1.28 & 86 & 12.1 & 1 & 0 & 0.975 & 1.87 & 80 & 22.3 & 1 & 0 & 0.972 & 3.87 & 86 & 13.9 & 1 & 0 & 0.974 & 2.4\\
    \textbf{SQReLU-QIP-1} & 438160 & \textbf{88} & \textbf{12.5} & \textbf{1} & \textbf{0} & \textbf{0.978} & \textbf{1.15} & \textbf{93} & \textbf{4} & \textbf{1} & \textbf{0} & \textbf{0.977} & \textbf{1.43} & 70 & 17.4 & 1 & 0 & 0.975 & 2.35 & \textbf{81} & \textbf{15.6} & \textbf{1} & \textbf{0} & \textbf{0.977} & \textbf{1.98}\\

    FQReLU-QFC-1-2 & 129168 & 84 & 14.3 & 1 & 0 & 0.971 & 4.14 & 76 & 13.7 & 1 & 0 & 0.971 & 1.21 & 84 & 16.9 & 1 & 0 & 0.972 & 1.66 & 87 & 15.4 & 1 & 0 & 0.971 & 2.55\\
    SQReLU-QFC-1-2 & 129168 & 82 & 15.1 & 1 & 0 & 0.973 & 2.49 & 73 & 17.9 & 1 & 0 & 0.973 & 2.59 & 88 & 9.9 & 1 & 0 & 0.975 & 3.32 & 90 & 4.9 & 1 & 0 & 0.974 & 1.7\\
    FQReLU-QIP-1-2 & 114776 & 89 & 14.7 & 1 & 0 & 0.972 & 1.41 & 84 & 19.2 & 1 & 0 & 0.97 & 5.75 & 78 & 17.6 & 1 & 0 & 0.966 & 2.7 & 84 & 14.8 & 1 & 0 & 0.967 & 3.77\\
    \textbf{SQReLU-QIP-1-2} & 114776 & \textbf{94} & \textbf{3.7} & \textbf{1} & \textbf{0} & \textbf{0.977} & \textbf{1.21} & {93} & {4.5} & {1} & {0} & {0.976} & {0.81} & 80 & 16.1 & 1 & 0 & 0.974 & 2.46 & 77 & 11.8 & 1 & 0 & 0.974 & 2.78\\

    FQReLU-QFC-1-4 & 34008 & 79 & 15.7 & 1 & 0 & 0.966 & 2.64 & 90 & 6.6 & 1 & 0 & 0.964 & 4.57 & 77 & 17.7 & 1 & 0 & 0.964 & 2.89 & 75 & 25 & 1 & 0 & 0.964 & 3.56\\
    SQReLU-QFC-1-4 & 34008 & 81 & 13.8 & 1 & 0 & 0.97 & 2.48 & 84 & 13.9 & 1 & 0 & 0.971 & 1.94 & 89 & 12 & 1 & 0 & 0.969 & 3.93 & 78 & 25.1 & 1 & 0 & 0.97 & 4.1\\
    FQReLU-QIP-1-4  & 27316 & 92 & 5.5 & 1 & 0 & 0.965 & 3.17 & 83 & 16.1 & 1 & 0 & 0.964 & 2.92 & 82 & 14.1 & 1 & 0 & 0.961 & 3.36 & 83 & 15.3 & 1 & 0 & 0.963 & 3.46\\
    SQReLU-QIP-1-4  & 27316 & 87 & 8.6 & 1 & 0 & 0.973 & 1.08 & 91 & 7.9 & 1 & 0 & 0.973 & 0.91 & 92 & 8.2 & 1 & 0 & 0.971 & 2.73 & 74 & 22.5 & 1 & 0 & 0.971 & 1.94\\

    \textbf{CONV-ReLU-IP} & 449000 & - & - & - & - & - & - & - & - & - & - & - & - & \textbf{98} & \textbf{1.1} & \textbf{1} & \textbf{0} & \textbf{0.98} & \textbf{0.49} & \textbf{91} & \textbf{13.8} & \textbf{1} & \textbf{0} & \textbf{0.98} & \textbf{0.86}\\
    \hline
    \end{tabular}
    }
    \caption{Performance of the models using different initialization methods for classifying images of the MNIST dataset. All models use quaternion convolution layers, except CONV-ReLU-IP.}
    \label{tab:qmnistperformance}
\end{table*} 

Some insights we obtained from this set of experiments are:
\begin{itemize}
    \item For the models with a higher number of parameters, there is no statistically significant difference between their performances regarding the selection of factors.
    \item SQReLU function achieves slightly better performance than FQReLU function.
    \item Quaternion-valued models achieve their best performance in fewer epochs than real-valued models.
    \item The type of initialization method affects in a statistically significant manner the performance of the models, being the best combinations for this dataset: Xavier initialization with QFC layers or Fully Quaternion initialization with QIP layers.
\end{itemize}

\subsection{CIFAR-10 dataset}

\begin{figure}[!t]
    \centering
    \includegraphics[width=0.375\textwidth]{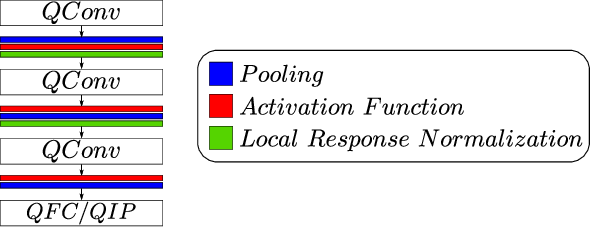} 
    \caption {Architecture used for classification of the CIFAR-10 dataset.} \label{fig:cifarmodels}
\end{figure}

The architecture used for this set of experiments is shown in Figure \ref{fig:cifarmodels}. Experiments were carried out in a similar way to the ones with the MNIST dataset. We tested the same 4 factors: initialization method, activation function, type of fully connected layer and number of parameters. Table ~\ref{tab:cifarmodels} shows the full list of models as well as their parameters for each layer. For each model, we obtained $8$ complete replicates, which consisted in training the model for $600$ epochs; data about the training and testing performance were saved for each $2$ epochs. Thereafter, for each replicate, the maximum performance value at the testing stage was selected, as well as the corresponding epoch and the training performance value at that epoch. This procedure was repeated for each model.

\begin{table*}
    \centering
    \begin{tabular}{|c|c|c|c|c|c|}
    \hline
    {Name} & {CONV-1} & {CONV-2} & {CONV-3} & {FC-1} & {No. Parameters} \\
    \hline
    FQReLU-QIP-1 & $16$ & $16$ & $32$ & $10$ & $79,680$ \\
    FQReLU-QIP-1-2 & $8$ & $8$ & $16$ & $10$ & $20,640$ \\
    FQReLU-QIP-1-4 & $4$ & $4$ & $8$ & $10$ & $5,520$ \\
    \hline
    FQReLU-QFC-1 & $16$ & $16$ & $32$ & $3$ & $78,784$ \\
    FQReLU-QFC-1-2 & $8$ & $8$ & $16$ & $3$ & $20,192$ \\
    FQReLU-QFC-1-4 & $4$ & $4$ & $8$ & $3$ & $5,296$ \\
    \hline
    SQReLU-QIP-1 & $16$ & $16$ & $32$ & $10$ & $79,680$ \\
    SQReLU-QIP-1-2 & $8$ & $8$ & $16$ & $10$ & $20,640$ \\
    SQReLU-QIP-1-4 & $4$ & $4$ & $8$ & $10$ & $5,520$\\
    \hline
    SQReLU-QFC-1 & $16$ & $16$ & $32$ & $3$ & $78,784$ \\
    SQReLU-QFC-1-2 & $8$ & $8$ & $16$ & $3$ & $20,192$ \\
    SQReLU-QFC-1-4 & $4$ & $4$ & $8$ & $3$ & $5,296$ \\
    \hline
    CONV-ReLU-IP & $32$ & $32$ & $64$ & $10$ & $80,640$ \\
     \hline
   \end{tabular}
   \caption{Models tested for classifying images of the CIFAR-10 dataset. The first $12$ rows show models that apply the quaternion convolution, while the last row is a model that applies the real-valued convolution (the name starts with CONV). Each model was named as follows: SQRELU indicates that it uses split quaternion activation functions; FQReLU indicates the use of fully quaternion activation functions, and ReLU indicates the use of real-valued ReLU functions. For fully connected layers: QIP indicates the use of quaternion inner product layers; QFC indicates the use of quaternion fully connected layers, and IP is used for real-valued inner product layers. The columns show the following information: CONV-1, CONV-2, CONV-3 indicate the number of outputs of each convolution layer (all kernels have sizes $5\times5$), if the output comes from a real-valued convolution layer, the value indicates the number of real-valued outputs, but if the output comes from a quaternion convolution layer, the value indicates the number of quaternion-valued outputs. FC-1 shows the number of outputs of the fully connected layer; in the same way, if the output comes from a quaternion-valued or real-valued inner product, the value indicates the number of real-valued outputs, but if the output comes from a quaternion fully connected layer, the value indicates the number of quaternion-valued outputs.} \label{tab:cifarmodels}
 \end{table*}

Using the selected data, a 4-way ANOVA test was conducted with the testing accuracy being the output variable. Normal population and equality of variances assumptions were checked. The analysis concluded that the overall interaction between the four factors was statistically significant ($F=5.521$, $p-value=4.751\times10^{-10}$); then, there is enough evidence supporting that the average accuracy of the groups is different. The 4-factor model is expressed by the following equation:

\begin{equation}
    Y_{ijkl} = \mu + \alpha_i+\beta_j+\gamma_k+\delta_l
    +(\alpha\beta\gamma\delta)_{ijkl}+\epsilon_{ijkl}
\end{equation}
where $Y_{ijkl}$ are the samples of the testing accuracy; $\mu$ is the grand mean; $\alpha_i=\mu-\mu_i$ is the effect of the initialization method, and $\mu_i$ is the mean of its population; $\beta_j=\mu-\mu_j$ is the effect of the activation function, and $\mu_j$ is the mean of its population; $\gamma_k=\mu-\mu_k$ is the effect of the activation function, and $\gamma_k$ is the mean of its population; $\delta_l=\mu-\mu_l$ is the effect of the activation function, and $\mu_l$ is the mean of its population; $(\alpha\beta\gamma\delta)_{ijkl}$ is the interaction effect of the initialization method, the activation function, the fully connected layer and the number of parameters; and $\epsilon_{ijkl}$ are the error terms, which are independently normally distributed random variables with expected value of zero.

\begin{figure*}[!t]
    \centering
    \includegraphics[width=\textwidth]{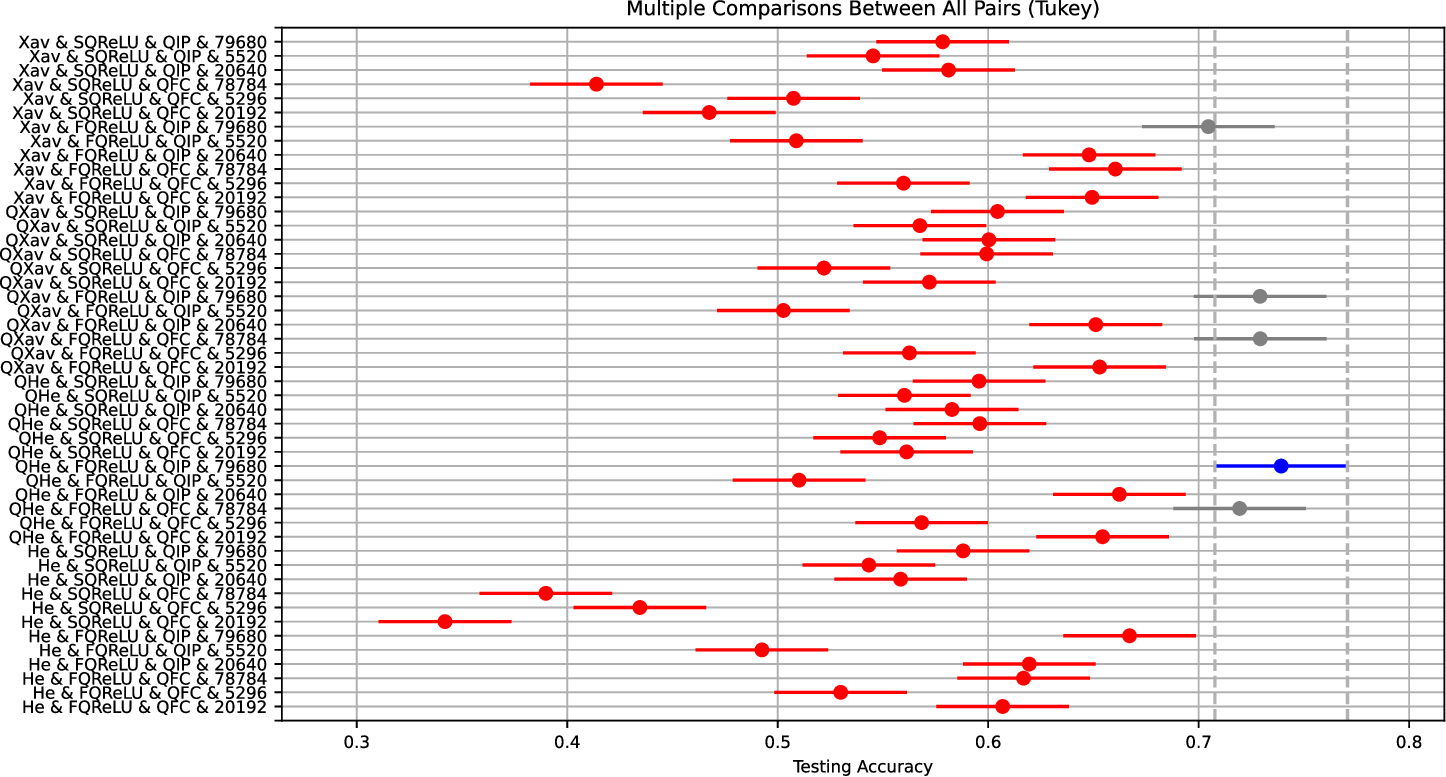}
    \caption {Plot of group means with confidence intervals using data from the CIFAR-10 classification task. The group with higher performance is selected in blue; the groups that are not statistically different from it are shown in gray, while the statistically different groups are shown in red.} \label{fig:cifarstats}
\end{figure*}

In order to determine the statistically significant combinations of parameters within each group, we conducted a Tukey Honestly Significant Difference (HSD) test, with $\alpha=0.05$. In this way, we found group combinations of the interaction terms which produce a superior effect on the testing accuracy (output variable). We found out that the best result was produced by the QHe initialization method, the FQReLU Activation Function, the QIP layer and the model with $79,680$ parameters (model FQReLU-QIP-1 in Table~\ref{tab:cifarmodels}). In addition, its performance was not statistically different from the one obtained by the QXavier and Xavier initialization methods ($meandiff=-0.0099$, $p-adjusted=0.9$ and $meandiff=-0.0349$, $p-adjusted=0.9$, respectively). In the same way, the model FQReLU-QFC-1, initialized with the QXavier or the QHe method, obtained similar and not statistically different performance ($meandiff=-0.0099$, $p-adjusted=0.9$ and $meandiff=-0.0197$, $p-adjusted=0.9$, respectively). The plot of group means with confidence intervals is shown in Figure \ref{fig:cifarstats}. 

From this analysis, we concluded that the FQReLU activation function outperforms the SQReLU activation function. In addition, when combined with a QIP layer, the model is more robust to the selection of the initialization method. A curious case occurs with model SQReLU-QFC-1 with Xavier and He initializations. Even though this model has the largest number of parameters, it obtained the worst performance of all the experiments, together with the FQReLU-QFC-1-4 model with He initialization.

\begin{figure*}[!t]
    \centering
    \includegraphics[width=0.485\textwidth]{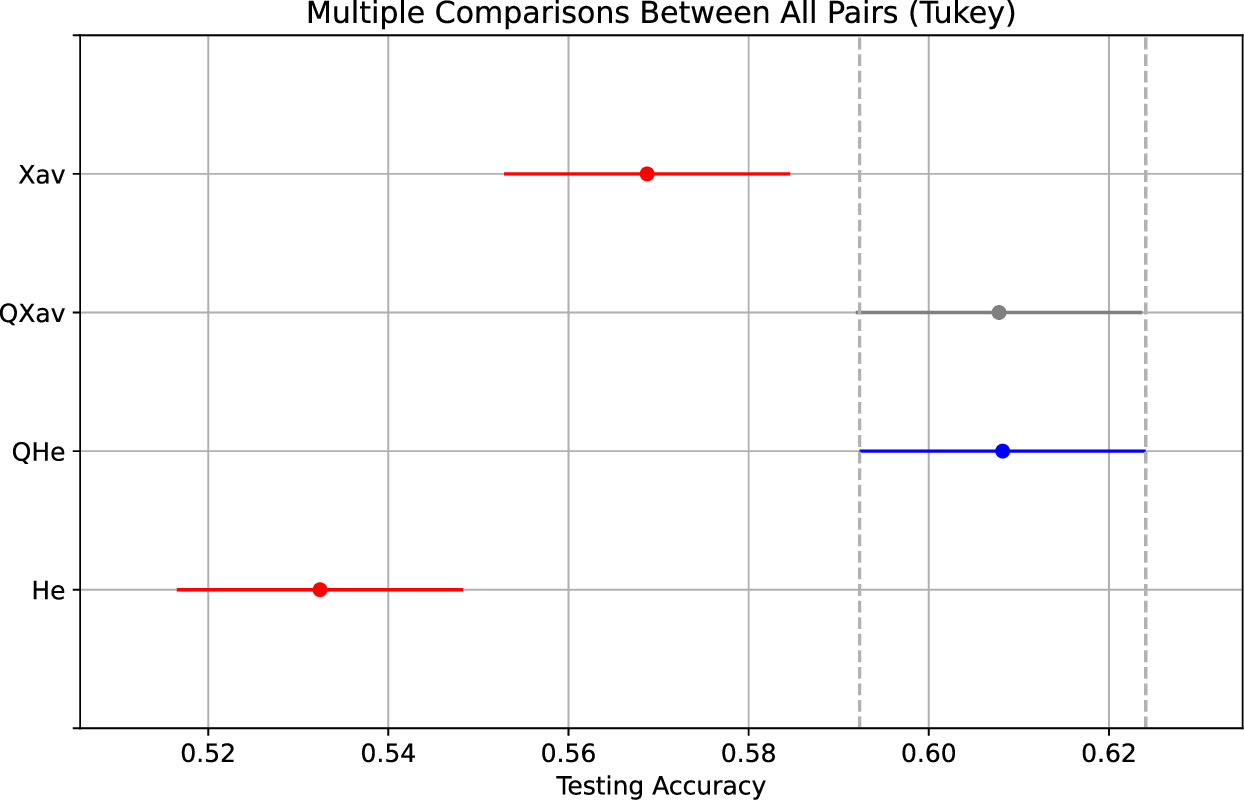} \hfill \includegraphics[width=0.495\textwidth]{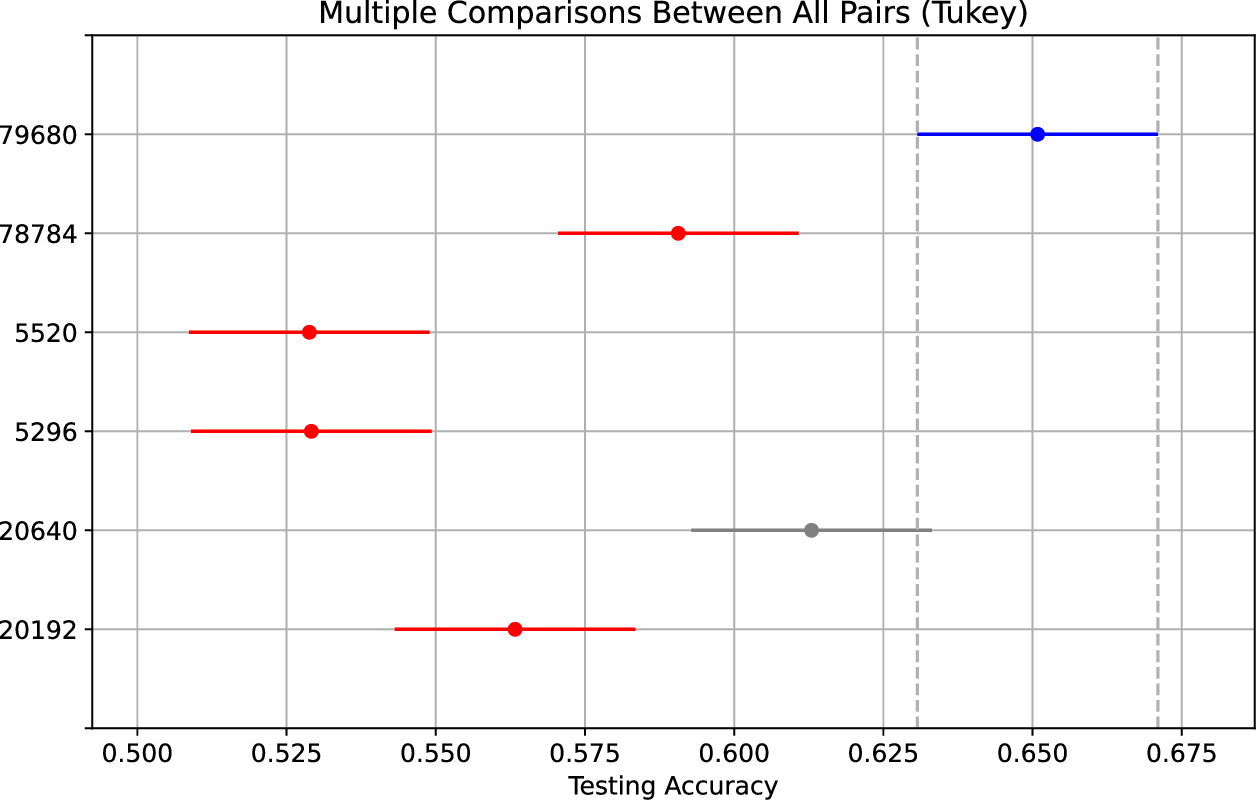}
    \caption {Plot of group means with confidence intervals using pooled data of factors from the CIFAR-10 classification task. Left: Initialization method. Right: Number of parameters. The group with higher performance is selected in blue; the groups that are not statistically different from it are shown in gray, while the statistically different groups are shown in red.} \label{fig:cifarfactors}
\end{figure*}

For the second analysis, we pooled the data of all factors except one, conducted a Tukey Honestly Significant Difference (HSD) test, with $\alpha=0.05$, and plotted the group means with confidence intervals, as is shown in Figure \ref{fig:cifarfactors}. We found out that for the activation function factor, FQReLU outperformed SQReLU function ($mean\ difference=0.0869$, $p-value=0.001$); for the fully connected layer, the QIP layer outperforms the QFC layer (mean difference$=0.0365$, $p-value=0.001$); and the initialization methods QXavier and QHe did not present a statistically significant difference in their performance and outperformed the real-valued versions (p-values included in the supplementary material). An interesting result was obtained when analyzing the number of parameters: when we pooled the data for this factor, the models FQReLU-QIP-1 and SQReLU-QIP-1 obtained the best performance, and it was not statistically different from the performance obtained by the pooled data of models FQReLU-QIP-1-2 and SQReLU-QIP-1-2 (p-values included in the supplementary material). Thus, the models with QIP layers can obtain similar results to the best model, but using a quarter of the number of parameters.

In a final analysis, we compared the quaternion models with best performance, i.e. FQReLU-QIP-1 (with QHe, QXavier and Xavier initializations) and FQReLU-QFC-1 (with QXavier and QHe initializations) versus real-valued models containing a similar number of parameters (CONV-ReLU-IP); we found out that there was no statistically significant difference in the performance between the real-valued model and the quaternion-valued models (p-values included in the supplementary material). In addition, all quaternion methods were trained in fewer epochs than the real-valued model with Xavier initialization. However, when the real-valued model used the He initialization, just $2$ of the quaternion-valued models were trained in fewer epochs: FQReLU-QFC-1 with QHe initialization and FQReLU-QIP-1 with QXavier initialization; where the former was the fastest model in the training stage. Finally, Table~\ref{tab:cifarperformance} shows the mean and standard deviation values of the training accuracy and the testing accuracy for each model. In addition, it is shown how many epochs it took, on average, to obtain the best performance of the model.

\begin{table*}
    \centering
    \resizebox{\textwidth}{!}{ 
    \begin{tabular}{|c|c|c|c|c|c|c|c|c|c|c|c|c|c|c|c|c|c|c|c|c|c|c|c|c|c|}
    \hline
    & & \multicolumn{6}{c|}{QXavier Initialization} & \multicolumn{6}{c|}{QHe Initialization} & \multicolumn{6}{c|}{Xavier Initialization} & \multicolumn{6}{c|}{He Initialization} \\
    \cline{3-26}
    {Name} & {No.} & \multicolumn{2}{c|}{Epochs} & \multicolumn{2}{c|}{$\%$ Train} & \multicolumn{2}{c|}{$\%$ Test} & \multicolumn{2}{c|}{Epochs} & \multicolumn{2}{c|}{$\%$ Train} & \multicolumn{2}{c|}{$\%$ Test} & \multicolumn{2}{c|}{Epochs} & \multicolumn{2}{c|}{$\%$ Train} & \multicolumn{2}{c|}{$\%$ Test} & \multicolumn{2}{c|}{Epochs} & \multicolumn{2}{c|}{$\%$ Train} & \multicolumn{2}{c|}{$\%$ Test} \\
    \cline{3-26}
    & Param & $\mu$ & $\sigma$ & $\mu$ & $\sigma$ & $\mu$ & $\sigma$ & $\mu$ & $\sigma$ & $\mu$ & $\sigma$ & $\mu$ & $\sigma$ & $\mu$ & $\sigma$ & $\mu$ & $\sigma$ & $\mu$ & $\sigma$ & $\mu$ & $\sigma$ & $\mu$ & $\sigma$ & $\mu$ & $\sigma$ \\
    \cline{3-26}
    &  & & & & $(10^{-2})$ & & $(10^{-2})$ & & & & $(10^{-2})$ & & $(10^{-2})$ & & & & $(10^{-2})$ & & $(10^{-2})$ & & & & $(10^{-2})$ & & $(10^{-2})$ \\
    \hline
    FQReLU-QFC-1  & 78784 & \textbf{479} & \textbf{83} & \textbf{0.828} & \textbf{2.12} & \textbf{0.729} & \textbf{1.01} & \textbf{399} & \textbf{133} & \textbf{0.826} & \textbf{1.3} & \textbf{0.72} & \textbf{1.71}& 458 & 167 & 0.756 & 4.75 & 0.66 & 4.19 & 421 & 81 & 0.688 & 3.65 & 0.617 & 2.62\\
    SQReLU-QFC-1  & 78784 & 458 & 51 & 0.694 & 1.69 & 0.599 & 0.91 & 460 & 58 & 0.695 & 2.45 & 0.596 & 1.33 & 466 & 89 & 0.485 & 4.96 & 0.414 & 2.54 & 555 & 30 & 0.44 & 8.12 & 0.39 & 6.43\\
    FQReLU-QIP-1 & 79680 & \textbf{415} & \textbf{85} & \textbf{0.816} & \textbf{2.13} & \textbf{0.729} & \textbf{1.04} & \textbf{473} & \textbf{61} & \textbf{0.83} & \textbf{1.31} & \textbf{0.739} & \textbf{0.7} & \textbf{478} & \textbf{123} & \textbf{0.796} & \textbf{3.07} & \textbf{0.705} & \textbf{1.39} & 439 & 116 & 0.755 & 2.62 & 0.667 & 2.02\\
    SQReLU-QIP-1 & 79680 & 444 & 84 & 0.675 & 3.16 & 0.604 & 0.96 & 395 & 106 & 0.688 & 2.66 & 0.596 & 1.7 & 448 & 158 & 0.656 & 2.88 & 0.578 & 2.6 & 493 & 84 & 0.656 & 4.07 & 0.588 & 2.14\\
    FQReLU-QFC-1-2 & 20192 & 401 & 144 & 0.729 & 1.96 & 0.653 & 1.6 & 343 & 111 & 0.736 & 2.72 & 0.654 & 1.83 & 455 & 93 & 0.729 & 1.25 & 0.649 & 1.74 & 386 & 104 & 0.706 & 2.67 & 0.607 & 2.07\\
    SQReLU-QFC-1-2 & 20192 & 436 & 163 & 0.643 & 2.82 & 0.572 & 2.42 & 389 & 161 & 0.641 & 2.85 & 0.561 & 1.62 & 470 & 129 & 0.499 & 7.53 & 0.467 & 7.08 & 391 & 190 & 0.391 & 10.56 & 0.342 & 10.43\\
    FQReLU-QIP-1-2 & 20640 & 418 & 143 & 0.729 & 2.23 & 0.651 & 1.33 & 381 & 136 & 0.749 & 1.46 & 0.662 & 2.02 & 418 & 121 & 0.726 & 2.83 & 0.648 & 1.77 & 524 & 91 & 0.711 & 2.23 & 0.62 & 1.83\\
    SQReLU-QIP-1-2 & 20640 & 549 & 40 & 0.694 & 4.14 & 0.6 & 2.78 & 398 & 164 & 0.659 & 2.95 & 0.583 & 3.42 & 464 & 117 & 0.656 & 3.42 & 0.581 & 1.95 & 466 & 122 & 0.639 & 2.23 & 0.558 & 2.29\\
    FQReLU-QFC-1-4 & 5296 & 410 & 162 & 0.634 & 4.1 & 0.563 & 3.04 & 430 & 166 & 0.626 & 3.29 & 0.568 & 1.41 & 446 & 73 & 0.635 & 3.38 & 0.56 & 1.11 & 466 & 64 & 0.606 & 4.34 & 0.53 & 3.11\\
    SQReLU-QFC-1-4 & 5296 & 354 & 114 & 0.589 & 3.27 & 0.522 & 2.24 & 438 & 82 & 0.611 & 5.11 & 0.548 & 1.86 & 544 & 52 & 0.551 & 3.64 & 0.508 & 2.55 & 506 & 102 & 0.475 & 7.52 & 0.435 & 8.26\\
    FQReLU-QIP-1-4 & 5520 & 350 & 120 & 0.577 & 2.66 & 0.503 & 2.17 & 314 & 149 & 0.586 & 2.2 & 0.51 & 2.03 & 441 & 91 & 0.586 & 3.07 & 0.509 & 2.12 & 410 & 140 & 0.559 & 3.4 & 0.493 & 3.36\\
    SQReLU-QIP-1-4 & 5520 & 466 & 90 & 0.669 & 3.09 & 0.568 & 1.88 & 404 & 143 & 0.643 & 2.31 & 0.56 & 2.56 & 328 & 140 & 0.63 & 2 & 0.545 & 2.09 & 521 & 85 & 0.613 & 4.68 & 0.543 & 2.38\\

    \textbf{CONV-ReLU-IP} & 80640 & - & - & - & - & - & - & - & - & - & - & - & - & 495 & 111 & 0.8 & 4.11 & 0.718 & 3.19 & \textbf{453} & \textbf{102} & \textbf{0.755} & \textbf{8.72} & \textbf{0.68} & \textbf{5.8}\\
    \hline
   \end{tabular}
   }
    \caption{Performance of the models using different initialization methods for classifying images of the CIFAR10 dataset. All models use quaternion convolution layers, except CONV-ReLU-IP.} \label{tab:cifarperformance}

 \end{table*} 

Some insights we obtained from this set of experiments are:
\begin{itemize}
    \item There exists an interaction effect between the atomic components of the models, which causes better or worse performance in this classification task.
    \item The best configurations for the CIFAR-10 dataset are:
    \begin{itemize}
        \item FQReLU activation function, any fully connected layer, and QHe or QXavier initialization.
        \item FQReLU activation function, QIP layer, and Xavier initialization method.
    \end{itemize}
    \item The FQReLU function achieves better performance than the SQReLU function.
    \item The QIP layer achieves better performance than the QFC layer.
    \item Fully quaternion initialization methods achieve better performance than channel-wise initialization ones.
    \item In some cases, quaternion models achieve their best performance in fewer epochs than real-valued models.
\end{itemize}

\section{Discussion} \label{sec:discussion}

\subsection{What is the effect of the independent components?}
When we analyzed the factors without their interaction effects, we found out that:
\begin{itemize}
    \item In models where the activation function is used extensively, the FQReLU function has better performance than the SQReLU function (an improvement in accuracy of $8.9\%$). See, for example, the model used for the CIFAR10 classification task, with $3$ activation function modules, versus the model used for the MNIST classification task, with just $1$ activation function module.
    \item In models where the fully connected module is used extensively, QIP or QFC modules achieve similar results; however when it only uses a fully connected module, the QIP module has better performance than the QFC one (an improvement in accuracy of $3.65\%$). Further analysis is required to test which of these modules is the best when it is used extensively in the architecture.
    \item For small models, like the one used for the MNIST classification task, there is no statistically significant difference between the initialization methods; however, for larger models, like the one used for the CIFAR classification task, fully quaternion initialization methods produced the best results.
\end{itemize}

\subsection{Can we take advantage of the interaction effect to design more compact models?}

We found out that the interaction effect of the $4$ tested factors is only visible in deeper models, e.g. the model used for the CIFAR dataset ($3$ convolution layers with $4$ factors having interaction) vs. the model used for the MNIST dataset ($2$ convolution layers with $2$ factors having interaction). 

In addition, when we took into account the interaction effects of the factors, we found out from the MNIST classification task that the interaction effects produce a better result when Xavier initialization is combined with QFC layers or QXavier initialization is combined with QIP layers, no matter the choice of the activation function. However, all of the models with the highest number of parameters produce not statistically different results. In addition, models with $4x$ fewer parameters have the same performance, with no statistically significant difference from the best models. These models have an interaction effect between the He or Xavier initialization and the QFC layer, or the QXavier or QHe initialization and the QIP layer. This clearly shows that the correct selection of factors can lead to more compact models with the same performance.

Now, if we consider the CIFAR dataset, the interaction effects produce a better result when we use FQReLU activation function, QFC or QIP layer, and QHe or QXavier initialization; or FQReLU activation function, QIP layer, and Xavier initialization method. However, when grouping data by the number of parameters, we found out that there is no statistically significant difference between the larger models and their corresponding models with $4x$ fewer parameters, see Figure \ref{fig:cifarfactors} (right).

\section{Conclusions}\label{sec:conclusions}

This paper contributes to the ongoing research on Quaternion-valued CNNs by addressing a key question in the field: how do different adaptations of components impact model performance? Having this question in mind, we presented the first statistical proof of the existence of interaction effects between different components of a Quaternion-valued CNN. In addition, our analysis provided the following insights:
\begin{itemize}
    \item Quaternion-valued CNN models can achieve similar results than the real-valued ones.
    \item The majority of QCNN models converge in fewer epochs than the real-valued models.
    \item The FQReLU activation function that we proposed produces better results than the SQReLU one ($8.9\%$ better at the CIFAR-10 dataset).
    \item Fully quaternion initialization methods outperform split quaternion ones.
    \item The interaction effects can improve the performance of some models in such a way that no statistically significant difference can be found between a large quaternion model and another with $4x$ fewer parameters.
\end{itemize}

Because of the No Free Lunch theorems~\cite{1995:wolpert:nofreelunch,wolpert:1997:nofreelunch}, one could say that our results might change with different problems. However, since QCNNs have been used in the context of image classification, we can infer that similar results would be obtained in this and similar domains. Also, in many cases, different combinations achieved a good performance, thus if a ``good enough’’ solution is sufficient, then the precise combination and parameters used will not be relevant. However, in critical applications that require the best available performance, according to these results, a practical insight when designing a model is to try the following combination of factors: FQReLU activation function, QHE or QXAVIER initialization methods (indistinctively), and test QIP and QFC layers.

Finally, this paper followed a systematic approach using statistical techniques to study the behaviour of QCNN architectures; following the same approach, future work should focus on comparing other components, such as channel-wise batch normalization~\cite{ioffe:2015:batchnormalization} versus quaternion batch normalization~\cite{gaudet:2018:qcnn,yin:2019:qcnn,grassucci:2022:qsngan}, testing different mappings of input data to the quaternion domain, comparing other quaternion activation functions available in the literature~\cite{parcollet:2018:qcnn, zhu:2018:qcnn, parcollet:2019:qcnnrecurrent, parcollet:2019:qcnn,grassucci:2021:qcnn} or proposing novel fully quaternion activation functions. In addition, this type of study should be extended to tasks in other domains, e.g. natural language processing, time series modeling, or generative tasks.

\section{CRediT authorship contribution statement}
\textbf{G. Altamirano:} Conceptualization, Data curation, Formal analysis, Funding acquisition, Investigation, Methodology, Software, Visualization and Writing-original draft.
\textbf{C. Gershenson:} Funding acquisition, Resources, Supervision and Writing-Review \& editing

\section{Declaration of Competing Interest}
The authors declare that they have no known competing financial interests or personal relationships that could have appeared to influence the work reported in this paper.

\section{Data Availability Statement}
Data will be made available on request.

\section{Acknowledgments}
G.Altamirano received a Postdoctoral Fellowship \textit{Estancias Posdoctorales por México} from CONACYT. C. Gershenson acknowledges support from UNAM-PAPIIT (IN107919, IV100120, IN105122), and the PASPA program from UNAM-DGAPA.

\bibliographystyle{elsarticle-num-names} 
\bibliography{bibliography}

\end{document}